\documentclass[sigconf]{acmart}

\usepackage{graphicx}
\usepackage{subcaption}
\usepackage{algorithm}
\usepackage{algorithmic}
\usepackage{multirow}
\usepackage{pgfplots}

\AtBeginDocument{%
  \providecommand\BibTeX{{%
    \normalfont B\kern-0.5em{\scshape i\kern-0.25em b}\kern-0.8em\TeX}}}

\copyrightyear{2020} 
\acmYear{2020} 
\setcopyright{acmlicensed}\acmConference[SIGIR '20]{Proceedings of the 43rd International ACM SIGIR Conference on Research and Development in Information Retrieval}{July 25--30, 2020}{Virtual Event, China}
\acmBooktitle{Proceedings of the 43rd International ACM SIGIR Conference on Research and Development in Information Retrieval (SIGIR '20), July 25--30, 2020, Virtual Event, China}
\acmPrice{15.00}
\acmDOI{10.1145/3397271.3401113}
\acmISBN{978-1-4503-8016-4/20/07}


\begin{document}

\title{Distributed Equivalent Substitution Training for Large-Scale Recommender Systems}

\author{Haidong Rong}
\email{hudsonrong@tencent.com}
\affiliation{\institution{Tencent Inc.}}

\author{Yangzihao Wang}
\email{slashwang@tencent.com}
\affiliation{\institution{Tencent Inc.}}

\author{Feihu Zhou}
\email{hopezhou@tencent.com}
\affiliation{\institution{Tencent Inc.}}

\author{Junjie Zhai}
\email{jasonzhai@tencent.com}
\affiliation{\institution{Tencent Inc.}}

\author{Haiyang Wu}
\email{gavinwu@tencent.com}
\affiliation{\institution{Tencent Inc.}}

\author{Rui Lan}
\email{franklan@tencent.com}
\affiliation{\institution{Tencent Inc.}}

\author{Fan Li}
\email{oppenheimli@tencent.com}
\affiliation{\institution{Tencent Inc.}}

\author{Han Zhang}
\email{lavenzhang@tencent.com}
\affiliation{\institution{Tencent Inc.}}

\author{Yuekui Yang}
\email{yuekuiyang@tencent.com}
\affiliation{\institution{Tencent Inc.}}

\author{Zhenyu Guo}
\email{alexguo@tencent.com}
\affiliation{\institution{Tencent Inc.}}

\author{Di Wang}
\email{diwang@tencent.com}
\affiliation{\institution{Tencent Inc.}}

\renewcommand{\shortauthors}{Haidong Rong et al.}

\fancyhead{}

\begin{abstract}
We present Distributed Equivalent Substitution (DES) training, a novel distributed training framework for large-scale recommender systems with dynamic sparse features. DES introduces fully synchronous training to large-scale recommendation system for the first time by reducing communication, thus making the training of commercial recommender systems converge faster and reach better CTR~\@. DES requires much less communication by substituting the weights-rich operators with the computationally equivalent sub-operators and aggregating partial results instead of transmitting the huge sparse weights directly through the network. Due to the use of synchronous training on large-scale Deep Learning Recommendation Models (DLRMs), DES achieves higher AUC(Area Under ROC). We successfully apply DES training on multiple popular DLRMs of industrial scenarios. Experiments show that our implementation outperforms the state-of-the-art PS-based training framework, achieving up to 68.7\% communication savings and higher throughput compared to other PS-based recommender systems.
\end{abstract}



\maketitle

\section{Introduction}
\label{sec:intro}
Large-scale recommender systems are critical tools to enhance user experience and promote sales/services for many online websites and mobile applications. One essential component in the recommender system pipeline is click-through rate (CTR) prediction. Usually, people use machine learning models with tens or even hundreds billions of parameters to provide the prediction based on tons of streaming input data that include user preferences, item features, user-item past interactions, etc. Current industrial-level recommender systems(RSs) usually have so large parameter size that asynchronous parameter-server (PS) mode has become the only available method for building such systems.

Ideally, an efficient distributed recommender system should meet three requirements:
\begin{itemize}
    \item \textbf{Dynamic Features:} In industrial scenarios, more and more recommender systems run on streaming mode because new users or items arrive continuously in infinite data streams. In the streaming recommender systems \cite{Gholami:2017:IMB,Chang:2017:SRS}, the size of model parameters is usually temporal dynamic and reaches hundreds of GBs or even several TBs. Such large-scale of the parameters naturally requires distributed storage.
    \item \textbf{Stable Convergence:} Before the popularity of DLRMs, the negative impacts on  accuracy caused by gradient staleness ~\cite{RDS:2016:Chen}in asynchronous training is not significantly in RSs. With more and more deep learning components are introduced to recommendation models, the RSs are required to supporting fully synchronization training for stable convergence and higher AUC~\@.
    \item \textbf{Real-time Updating:} One vital characteristic of streaming recommendation scenarios is their high velocity of inference query. So an RS needs to update and response instantly in order to catch users' real-time intention and demands. With model size increasing over time, it is more and more important for RSs to reduce the demand of network transmission to keep timeliness.
\end{itemize}

The above requirements are affected by two design choices we make when building a large-scale distributed recommender system: how to parallelize the training pipeline, and how to synchronize the parameters. For parallelization, we can use either data parallelism (to parallelize over the data dimension), or model parallelism (to parallelize computation on parameters on different devices). For synchronization, the system can be synchronous or asynchronous (usually when using PS mode).

However, existing methods cannot be easily adapted to recommender systems for two reasons:

First, for the DLRMs with very large size of parameters, pure data parallelism keeps replica of the entire model on a single device , which makes it impossible because recommender systems usually have very large weights to updating for the first few layers (we call operators in these layers \textit{weights-rich layers}). Also, in the context of recommender system, features for different input samples in a batch can be different in length, so pure data parallelism with linearly-scaled batch size is inapplicable. Pure model parallelism usually treat the layers and operators as a whole and optimize the load balance by different device placement policies, which does not apply to most larger-scale recommender systems today either.

Second, current PS mode implementations of large-scale recommender systems is essentially a hybrid-data-and-model parallelism strategy and always needs to make a tradeoff between update frequency and communication bandwidth. Applying such asynchronous strategy to current and future models with even larger size of parameters will make it more difficult for these models to converge to the same performance while keeping the training efficient.

To solve the above two issues, we present a novel distributed training framework for recommender systems that achieves faster training speed with less communication overhead using a strategy we call \textit{distributed equivalent substitution (DES)}. The key idea of DES is to replace the weights-rich layers by an elaborate group of sub-operators which make each sub-operator only update its co-located partial weights. The partial computation results get aggregated and form a computationally equivalent substitution to the original operator. To achieve less communication, we find sub-operators that generate partial results with smaller sizes to form the equivalent substitution. We empirically show that for all the weights-rich operators whose parameters dominate the model, it is easy to find an equivalent substitution strategy to create an order of magnitude less communication demand. We also discuss how to extend DES to other general models\footnote{More details in Section~\ref{sec:app}.}.

The main contributions of this paper are as follows:
\begin{itemize}
    \item We present DES training, a distributed training method for recommender systems that achieves better convergence with less communication overhead on large-scale streaming recommendation scenarios.
    \item We propose a group of strategies that replaces the weights-rich layers in multiple popular recommendation models by computationally equivalent sub-operators which only update co-located weights and aggregate partial results with much smaller communication cost.
    \item We show that for different types of models that are most often used in recommender systems, we can find according substitution strategies for all of their weights-rich layers.
    \item We present an implementation of DES training framework that outperforms the state-of-the-art recommender system. In particular, we show that our framework achieves 68.7\% communication savings on average compared to other PS-based recommender systems.
\end{itemize}

\section{Related Work}
\label{sec:related_work}
Large-scale recommender systems are distributed systems designed specifically for training recommendation models. This section reviews related works from the perspectives of both fields:
\subsection{Large-Scale Distributed Training Systems}
\textbf{Data Parallelism} splits training data on the batch domain and keeps replica of the entire model on each device. The popularity of ring-based AllReduce~\cite{Gibiansky:2017:BHT} has enabled large-scale data parallelism training~\cite{Goyal:2017:ALM, Jia:2018:HSD, You:2019:LBO}. \textbf{Parameter Server} (PS) is a primary method for training large-scale recommender systems due to its simplicity and scalability~\cite{Dean:2012:LSD, Li:2014:SDM}. Each worker processes on a subset of the input data, and is allowed to use stale weights and update either its weights or that of a parameter server. \textbf{Model Parallelism} is another commonly used distributed training strategy~\cite{Krizhevsky:2014:OWT, Dean:2012:LSD}. More recent model parallelism strategy learns the device placement~\cite{Mirhoseini:2017:DPO} or uses pipelining~\cite{Huang:2018:GET}. These works usually focus on enabling the system to process complex models with large amount of weights.

Previously, there have been several hybrid-data-and-model parallelism strategies. Krizhevsky~\cite{Krizhevsky:2014:OWT} proposed a general method for using both data and model parallelism for convolutional neural networks. Gholami et al.~\cite{Gholami:2017:IMB} developed an integrated model, data, and domain parallelism strategy. Though theoretically summarized several possible ways to distribute the training process, the method only focused on limited operations such as convolution, and is not applicable to fully connected layers. Zhihao et al.~\cite{Jia:2018:EHD} proposed another integrated parallelism strategy called "\textit{layer parallelism}". However, it also focuses on a limited set of operations and cannot split the computation for an operation, which makes it difficult to apply this method to recommender systems. Mesh-TensorFlow~\cite{Shazeer:2018:MDL} implements a more flexible parameter server-like architecture, but for recommender systems, it could introduce unnecessary weights communication between different operations.

\subsection{Recommender Systems}
 The critical problem a recommender system tries to solve is the Click-Through Rate (CTR) prediction. Logistic regression (LR) is one of the first methods that has been applied~\cite{Richardson:2007:PCE} and is still a common practice now. Factorization machine (FM)~\cite{Rendle:2010:FM} utilizes addition and inner product operations to capture the linear and pairwise interactions between features. More recently, deep-learning based recommendation models(DLRMs) have gained more and more attentions~\cite{Zhang:2016:DLO, Cheng:2016:WDL, Guo:2017:DFB, Lian:2018:XCE, Zhou:2018:DIN}. Wide \& Deep(W\&D)  model combines a general linear model (the wide part) with a deep learning component (the deep part) to enable the recommender to capture both memorization and generalization. DeepFM seamlessly integrates factorization machine and multi-layer perceptron (MLP) to model both the high-order and low-order feature interactions. Other applications of DLRM include music recommendation~\cite{Oord:2013:DCM} and video recommendation~\cite{Covington:2016:DNN}. Among all the existing industrial-level recommender systems, one common characteristic is tens or even hundreds billions of dynamic features. To the best knowledge of the authors, the dominant way to build a large-scale recommender system today is still parameter-server based methods.

\section{Background and Design Methodology}
\label{sec:background}
\subsection{Recommender System Overview}

The typical process of a recommender system starts when a user-generated query comes in. The recommender system will return a list of items for the user to further interact (clicking or purchasing) or ignore. These user operations, queries and interactions are recorded in the log as training data for future use. Due to the large number of simultaneous queries in recommender systems, it is difficult to score each query in detail within the service latency requirement (usually 100 milliseconds). Therefore, we need a recall system to pick from the global item list a most-relevant short list, using a combination of machine learning models and manually defined rules. After reducing the candidate pool, a ranking system ranks all items according to their scores. The score $P$ usually presents the probability of user behavior tag $y$ for a given feature $x$ includes user characteristics (e.g., country, language, demographic), context features (e.g., devices, hours of the day, days of the week) and impression features (e.g., application age, application history statistics). This paper mainly studies the core component of a recommender system: models that are used for ranking and online learning.

\subsection{Distributed Equivalent Substitution Strategy}
Previous PS-based or model parallelism methods usually do not change the operator on algorithm level. That means for recommender systems that have weights-rich layers for the first one or more layers, putting operators on different devices still cannot solve the out-of-memory problem for a single weights-rich layer. Some works do split the operator~\cite{Huang:2018:GET, Jia:2018:EHD}, but they focus on the convolution, which has completely different characteristics than operators that are frequently used in recommender systems. Our strategy, instead, designs a computationally equivalent substitution for the original weights-rich layer, replace it into a group of computational equivalent operators that update only portions of weights, and processes the computation on non-overlapping input data. Since only one portion of weights is updated by one of new operators, our method could break through the single-node memory limitation and avoid transmitting a large number of parameters between the nodes. This strategy is particularly designed for large-scale recommender systems. In models for such recommender systems, the majority of the parameters only participate in very simple computation in the first few layers. Such models include LR, FM, W\&D, and many other follow-ups.

\subsubsection{Definitions and Notations}
To help readers better follow our contributions in later sections, we hereby list some basic definitions and notations in the context of distributed training framework for recommender system.
We first define the $\bigoplus$ operation for the convenience of description:
\begin{equation}
    R = \bigoplus_{i=1}^N{r_i}
    \label{eq:agg}
\end{equation}
In the context of this paper, $\bigoplus$ is one of the MPI-style collective operations: $\bigoplus \in (AllReduce, AllGather)$. However, it can be any communicative-associative aggregation operation. $r_i$ presents local values hold by processor $i$, $R$ presents the final result. The following are some definitions we need for the description of DES strategy:
\begin{itemize}
    \item $F$: the original operator function;
    \item $\mathcal{M}$: the sub-operator function;
    \item $\mathcal{F}$: the computationally equivalent substitution of $F$;
    \item $f$: the local result for one substitution operator of $F$;
    \item $B$: batch size of samples on each iteration;
    \item $N$: number of worker processes;
    \item $m$: number of sub-operators;
    \item $X$: input tensor of an operator;
    \item $W, V$: weights tensor of an operator;
    \item $\alpha$: latency of the network.
    \item $C$: network bandwidth;
    \item $S_{f,w,g,\mathcal{M}}$: size of features, weights, gradients, or intermediate results in bytes;
\end{itemize}
Without losing generality, we suppose that each worker only has one process, so the number of workers is equal to the number of processes. We also assume that all operators only take one input tensor $\mathbf{X}$ and one weights tensor $\mathbf{W}$.

\subsubsection{Algorithm}

The key observation is that for models in recommender systems, there is always one or more weights-rich layers with dominant portion of the parameters. The core idea of DES strategy is to find a computationally equivalent substitution to the operator of these weights-rich layers, and to find a splitting method to reduce the communication among all the sub-operators.

\begin{figure}[h]
	\centering
	\includegraphics[width=0.4\textwidth]{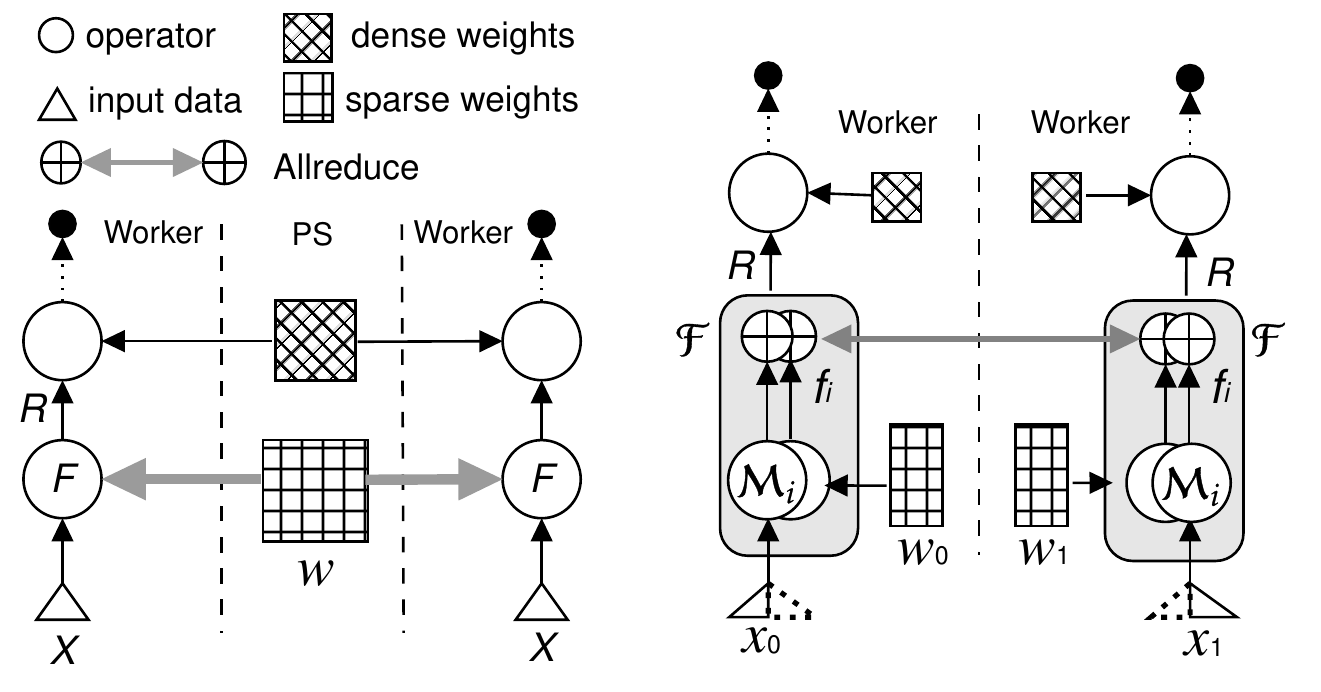}
	\caption{Forward pass for one operator of PS/Mesh-based strategy (left) and DES strategy (right).}
	\label{fig:des}
\end{figure}

\textbf{Forward Phase:} Figure~\ref{fig:des} illustrates the forward pass in two-worker case, and compares our DES strategy with PS-based strategy. In PS-based strategy, $F$ is not split, so each operator needs its entire $\mathbf{W}$ when doing the computation. Also, $\mathbf{W}$ is not co-located with $F$ but pulled to the device when needed. In DES strategy, we partition the weights and inputs on different processes, do parallel aggregations on results of one or more sub-operators $\{\mathcal{M}_i\}_{i=1}^m$, then use the substitution operator $\mathcal{F}$ to get the final result on each process. Algorithm~\ref{alg:des} shows this process:
\begin{algorithm}[htbp]
   \caption{Distributed Equivalent Substitution Algorithm}
   \label{alg:des}
\begin{algorithmic}
   \STATE {\bfseries Input:} data $\mathbf{X}$, weights $\mathbf{W}$, number of processes $N$, number of sub-ops $m$
   \REPEAT
   \STATE $\left\{W_{i}\right\}_{i=1}^{N} := \text{GetPartition}\left(\mathbf{W},N\right)$ 
   \STATE $\left\{X_{i}\right\}_{i=1}^{N} := \text{GetPartition}\left(\mathbf{X},N\right)$
   \STATE $\left\{\mathcal{M}_{j}\right\}_{j=1}^{m}, \mathcal{F} := \text{GetSubOperators}\left(F\right)$
   \STATE where $\mathcal{F}
   \left(\left\{\oplus\left(\mathcal{M}_{i}\left(W_i,X_i\right)\right)\right\}_{j=1}^{m}\right)\equiv F$ 
   \UNTIL{$\sum^m_{j=1}(S_{\mathcal{M}_j}) \ll S_W$}
   \FORALL{$i$-th process such that $1\leq i\leq N$}
   \STATE make $W_i$ and $X_i$ co-located with $i$-th process
   \FOR{$j=1$ {\bfseries to} $m$} 
   \STATE $f_j = \bigoplus_{j=1}^N \big(\mathcal{M}_j(w_i, x_i)\big)$ \COMMENT{parallel aggregation}
   \ENDFOR
   \STATE $R = \mathcal{F}(f_1, ..., f_{m})$
   \ENDFOR
   \RETURN $R$ \COMMENT{each process gets the same final results}
\end{algorithmic}
\end{algorithm}

The layers follow the weights-rich layer will get the same aggregated results on each process, so there is no need for further inter-process communication in subsequent computation for the forward phase. To guarantee the correctness of equation~\ref{alg:des}, it is very important that $\mathcal{F}$ is computationally equivalent to the original operator $F$. We observe that on all the popular models for recommender systems, we can always find such sub-operators to form computational equivalent substitutions. We will show details on how we get the substitutions for operators in different models in section~\ref{sec:app}.

\textbf{Back-propagation Phase:}After the forward phase, each process has the entire results $R$. Because we are not doing AllReduce on the gradients, but only on some small intermediate results, and also because aggregation operation distributes gradients equally to all its inputs, there is no inter-process communication during the back-propagation phase either. Each process just transfers the gradients directly back to its own sub-operator.

\subsubsection{Performance \& Complexity Analysis}

\textbf{PS-based:} Weights are distributed on parameter-servers, while $N$ workers process on $N$ different batches each with $B$ samples. The time cost for PS-based mode is:
\begin{equation*}
    \begin{aligned}
    T_{sync,ps} &= 2N\Big(\alpha + \frac{B (S_{f}+S_{w})}{C}\Big) \\
    T_{async,ps} &= 2\Big(\alpha + \frac{B (S_{f}+S_{w})}{C}\Big)
    \end{aligned}
\end{equation*}

\textbf{Mesh-based:} A special form of PS-based is Mesh-based in which the weights are divided into $n$ chunks and co-located with some workers. It has smaller network cost than original PS-based strategies. In this strategy, each worker processes one batch, the time cost for $n$ batches in synchronous mode is:
\begin{equation*}
    \begin{aligned}
    T_{sync,mesh} = 2N\Big(\alpha + \frac{(N-1)B (S_{f}+S_{w})}{C}\Big) \\
    T_{async,mesh} = 2\Big(\alpha + \frac{(N-1)B (S_{f}+S_{w})}{C}\Big)
    \end{aligned}
\end{equation*}

\textbf{AllReduce:} A full replica of weights is stored on each worker. The workers synchronize the gradients every iteration. We use Ring-based AllReduce, the most widely-adopted AllReduce algorithm, as the default algorithm for the scope of this paper. The time cost of the communication is:
\begin{equation*}
    T_{ring} = 2(N-1)(\alpha + \frac{S_{g}}{NC})
\end{equation*}
Where $S_{g}$ is the size of gradients for the model.

\textbf{DES:} Each aggregation operation uses AllReduce, DES may use several such aggregation operations to form the final result, so the time cost of the communication is:
\begin{equation*}
    T_{DES} = \sum^m_{i=1} T_{ring}(S_{\mathcal{M}_j})
\end{equation*}
Where $m$ is the number of aggregation operations, and $S_{\mathcal{M}_j}$ is the size of intermediate results for the $j$th operation $\mathcal{M}_j$.
Let
\begin{equation*}
\mathcal{M}_{j} : W_{i} \rightarrow \mathbb{R}^{S}, S=S_{\mathcal{M}_{j}}
\end{equation*}
and we can see if $S \ll |W_i|$ is satisfied for each $\mathcal{M}_j$, DES will reduce communication cost.

For both PS-mode strategy, time complexity of the communication is proportional to batch size $B$. For AllReduce and DES-based strategies, time complexity of the communication is constant (because the number of aggregation operations is usually smaller than 3).

The benefits of DES strategy is three-fold: first, with new operators and their co-located weights, one can split an operator with a huge amount of weights into sub-operators with arbitrarily small amount of parameters, given abundant number of workers. This enables better scalability for our framework when compared to traditional PS-based frameworks; second, DES strategy does not send weights but instead intermediate results from sub-operators, which can be much smaller in size compared to the original weights. This can significantly reduce the total amount of communication needed for our framework; third, with the above two improvements, our framework brings synchronous training to large-scale recommender system. With fully-synchronization per-iteration, the model converges faster, which makes the training process more efficient.

\section{Applications on Models for Recommender Systems}
\label{sec:app}
We observe that many models in recommender systems share similar components (Table~\ref{table:common}). For example, LR model is the linear part of W\&D model; almost all models include first-order feature crossover; all FM-based models include second-order feature crossover; the deep component of W\&D model and DeepFM model share similar structures. An optimal DES strategy finds substitutions of first-order, second-order, or higher-order operations, which are usually simple computation but with a large number of weights. The goal is to achieve the same computation but with much smaller communication cost for sending partial results over the network. In this section, we describe how to find such computational equivalent substitutions for different models.

\begin{table}[t]
\caption{Some common components that are shared among different recommender system models.}
\label{table:common}
\vskip 0.15in
\begin{center}
\begin{small}
\begin{sc}
\begin{tabular}{lccr}
\toprule
Model & first-order & second-order & high-order \\
\midrule
LR    & \checkmark  &              &             \\
W\&D  & \checkmark  &              & \checkmark  \\
FM    & \checkmark  & \checkmark   &             \\
DeepFM& \checkmark  & \checkmark   & \checkmark  \\
\bottomrule
\end{tabular}
\end{sc}
\end{small}
\end{center}
\vskip -0.1in
\end{table}

\subsection{Logistic Regression}
Logistic Regression(LR)~\cite{Richardson:2007:PCE} is a generalized linear model that is widely used in recommender systems. Due to its simplicity, scalability, and interpretability, LR can be used not only as an independent model, but also an important component in many DLRMs, such as Wide\&Deep and DeepFM~\@. The form of LR is as follows: 
$$
F_{l r}(\mathbf{W}, \mathbf{X}) =\sigma\left(\mathbf{W}^{T} \mathbf{X}+b\right)
$$
where, $\mathbf{X} = [x_1, x_2, ..., x_d]$ and $\mathbf{W} = [w_1, w_2, ..., w_d]$ are two d-dimension vectors represent inputs and weights respectively, $b$ is the bias, and $\sigma(\cdot)$ is a non-linear transform, usually a sigmoid function for LR\@.  The major part of the computation in $F_{lr}$ is dot product. It is easy for us to find an $N\text{-partition}$ of $\mathbf{W}$: $\mathbf{W}=\bigcup_{i=1}^{N} W_{i}$, where $W_{i}$ denotes the subset of $\mathbf{W}$ co-located with the $i$-th process. 
We then define a local operator $\mathcal{M}_1$ on $W_i$:
\begin{equation}
\begin{aligned}
    \mathcal{M}_1\left(W_i\right)&=\sum_{\forall w_{j} \in W_{i}} w_{j} x_{j} \\ 
    \label{eq:lr_m1}
\end{aligned}
\end{equation}
We have the equivalent substitution $f_n^{lr}$ of $F_{lr}$:
\begin{equation}
\begin{aligned}
    f_i^{lr} &= \sigma\left(\oplus \mathcal{M}_1\left(W_i\right)+b\right) \\
    &= \sigma\left(\oplus \left(\sum_{\forall w_j \in W_i} {w_j*x_j}\right)+b\right)
	\label{equation:des_lr}
\end{aligned}
\end{equation}

\begin{figure}[htbp]
	\centering
	\includegraphics[width=0.5\textwidth]{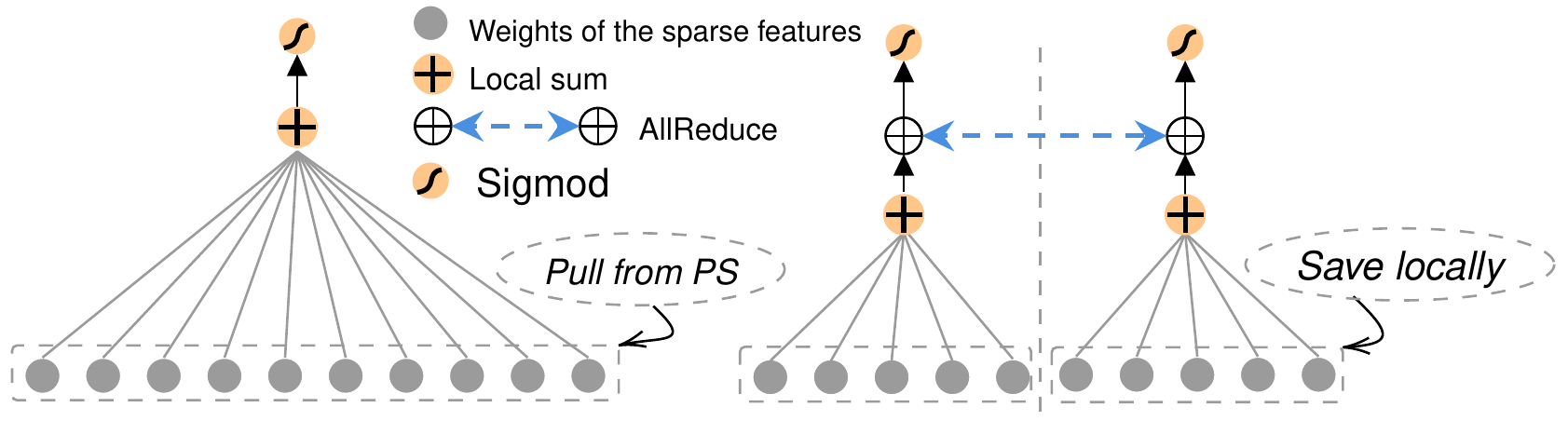}
	\caption{Forward pass for LR operator in PS/mesh-based strategy (left) and DES strategy when N=2 (right).}
	\label{fig:des_lr}
\end{figure}

Assume that all weights of sparse features are stored in hash tables as float32. In mesh-based strategy, each worker needs to transfer $\frac{N-1}{N}$ weights with unsigned int64 keys from the hash tables co-located with other workers. So the total data size to transfer through the network for each worker is:
$$Q_{lr}^{Mesh}=\frac{(N-1)}{N} \left(S_{f}+S_{w}\right)$$
Where $S_f$ and $S_w$ denote the size of feature keys and weights respectively.

Using DES, we only need to synchronize a scalar value with other workers for every sample, so the total data size to transfer through the network for each worker is:
$$Q_{lr}^{DES}=2\frac{(N-1)}{N}S_{\mathcal{M}_1}$$
Where $S_{\mathcal{M}_i}$ denotes the size of intermediate results.So the communication-saving ratio for LR is:
$$\mathcal{R}_{lr}=1-\frac{Q_{lr}^{DES}}{Q_{lr}^{Mesh}}=1-\frac{2S_{\mathcal{M}_1}}{S_k+S_w}$$

\subsection{Factorization Machine}
Besides linear interactions among features, FM models pairwise feature interactions as inner product of latent vectors. FM is both an independent model and an important component of DLRMs such as DeepFM and xDeepFM~\cite{Lian:2018:XDEEPFM}. The linear interactions are similar to LR model, so here we only focus on the order-2 operator (denoted by $fm(2)$):
\begin{equation}
\begin{aligned}
    F_{fm(2)}=&\sum_{i=1}^{d} \sum_{j=i+1}^{d}\left\langle v_{i}, v_{j}\right\rangle x_i \cdot x_{j} \\
    =&\frac{1}{2} \sum_{i=1}^{d} \sum_{j=1}^{d}\left\langle{v}_{i}, {v}_{j}\right\rangle x_{i} x_{j}-\frac{1}{2} \sum_{i=1}^{d}\left\langle{v}_{i}, {v}_{i}\right\rangle x_{i} x_{i} \\
    =&\frac{1}{2}\left\langle\sum_{i=1}^{d} {v}_{i} x_{i}, \sum_{i=1}^{d} {v}_{i} x_{i}\right\rangle-\frac{1}{2}\sum_{i=1}^{d}\left\langle{v}_{i} x_{i}, {v}_{i} x_{i}\right\rangle
    \label{eq:fm_linear}
\end{aligned}
\end{equation}
$v_{i}$ denotes a latent vector, $x_i$ is the feature value of $v_{i}$, the $\left\langle\cdot\right\rangle$ presents the inner product operation.

Equation~\ref{eq:fm_linear} shows another popular form for FM mentioned in ~\cite{Rendle:2010:FM} with only linear complexity. Here we adopt this equation to form our computational equivalent substitution of FM~\@.

Applying Algorithm~\ref{alg:des} to FM, we get an $N$-partition of $\mathbf{V}=\bigcup_{i=1}^{N} V_{i}$ using any partition policy that balances $|V_i|$ on each process. We then define two local operators: $\mathcal{M}_1$ and $\mathcal{M}_2$ that process on local subset of weights $V_i$:
\begin{equation}
\begin{aligned}
    \mathcal{M}_1\left(V_i\right)&=\sum_{\forall v_{j} \in V_{i}} v_{j} x_{i} \\ 
    \mathcal{M}_2\left(V_i\right)&=\sum_{\forall v_{j} \in V_{i}}\left\langle v_{j} x_{j}, v_{j} x_{j}\right\rangle
    \label{eq:fm_m1m2}
\end{aligned}
\end{equation}
We have the equivalent substitution $f_i^{fm(2)}$ of $F_{fm(2)}$:

\begin{equation}
\begin{aligned}
    f_i^{fm(2)} =\frac{1}{2}\left\langle\oplus\mathcal{M}_1\left(V_i\right),\oplus\mathcal{M}_1\left(V_i\right)\right\rangle-\frac{1}{2}\oplus\mathcal{M}_2(V_i)
    \label{eq:des_fm2}
\end{aligned}
\end{equation}

\begin{figure}[htbp]
	\centering
	\includegraphics[width=0.5\textwidth]{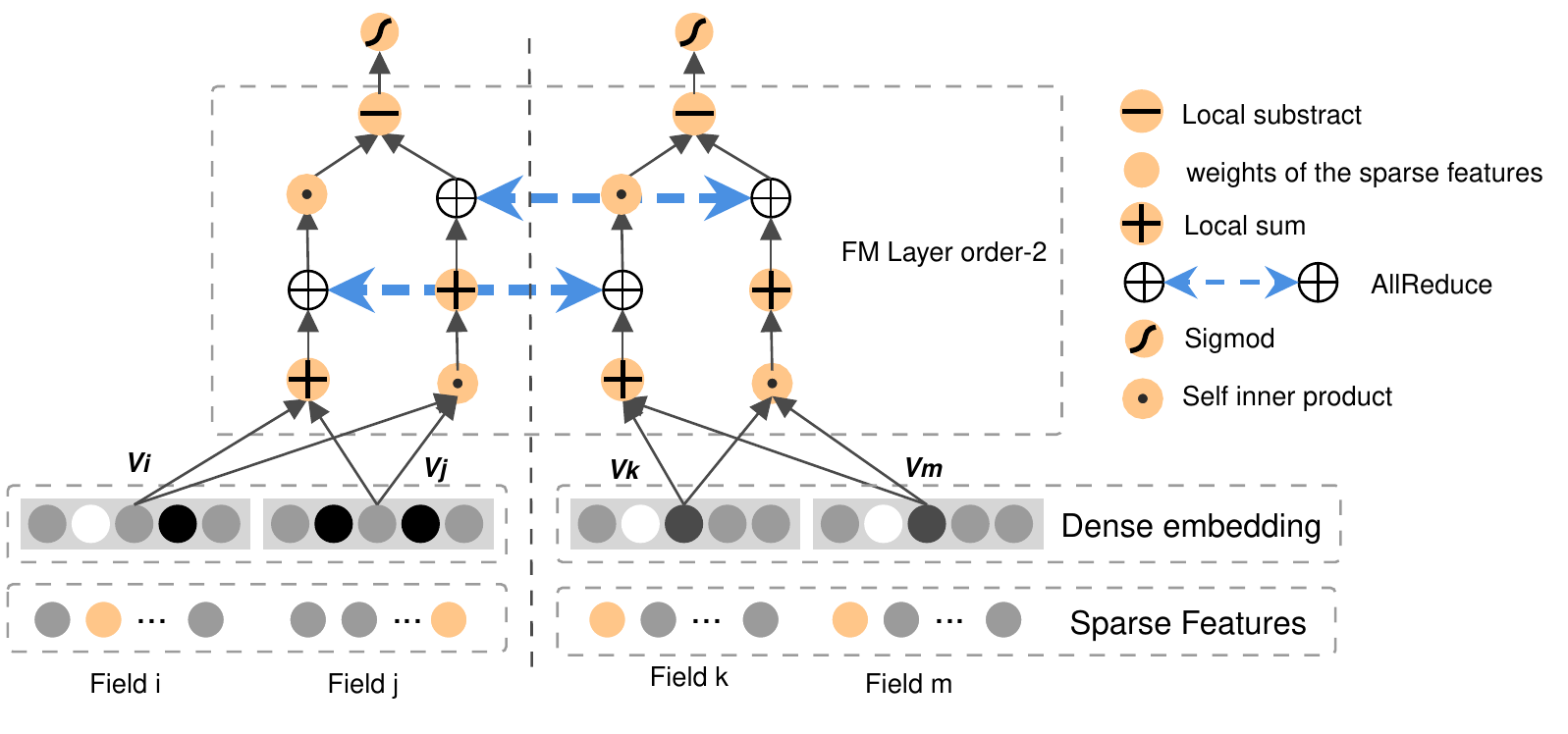}
	\caption{Forward pass for FM order-2 operators using DES strategy when $N$=2.}
	\label{fig:fm_impl}
\end{figure}

In mesh-based strategy, each worker needs to lookup $\frac{N-1}{N}$ latent vectors with feature IDs from the hash tables co-located with other workers. The total data size to transfer through the network for each worker is:
$$Q_{fm(2)}^{Mesh}=\frac{(N-1)}{N} \left(S_{f}+S_{V}\right)$$
Where $S_f$ and $S_V$ denote the size of feature keys and latent vectors per batch respectively.

Using DES, the FM order-2 operators only require all workers to exchange $\mathcal{M}_1(V_i)$ and $\mathcal{M}_2(V_i)$ among each other, so we have:
$$Q_{fm(2)}^{DES}=2\frac{(N-1)}{N} \left(S_{\mathcal{M}_1}+S_{\mathcal{M}_2}\right)$$

The communication-saving ratio for FM is:
$$\mathcal{R}_{fm(2)}=1-\frac{Q_{fm(2)}^{DES}}{Q_{fm(2)}^{Mesh}}=1-\frac{2\left(S_{\mathcal{M}_1}+S_{\mathcal{M}_2}\right)}{S_f+S_V}$$

\subsection{Deep Neural Network}
\begin{figure}[htbp]
	\centering
	\includegraphics[width=0.17\textwidth]{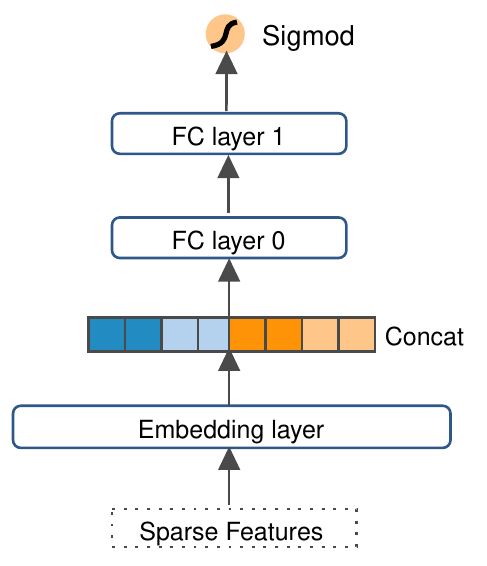}
	\includegraphics[width=0.26\textwidth]{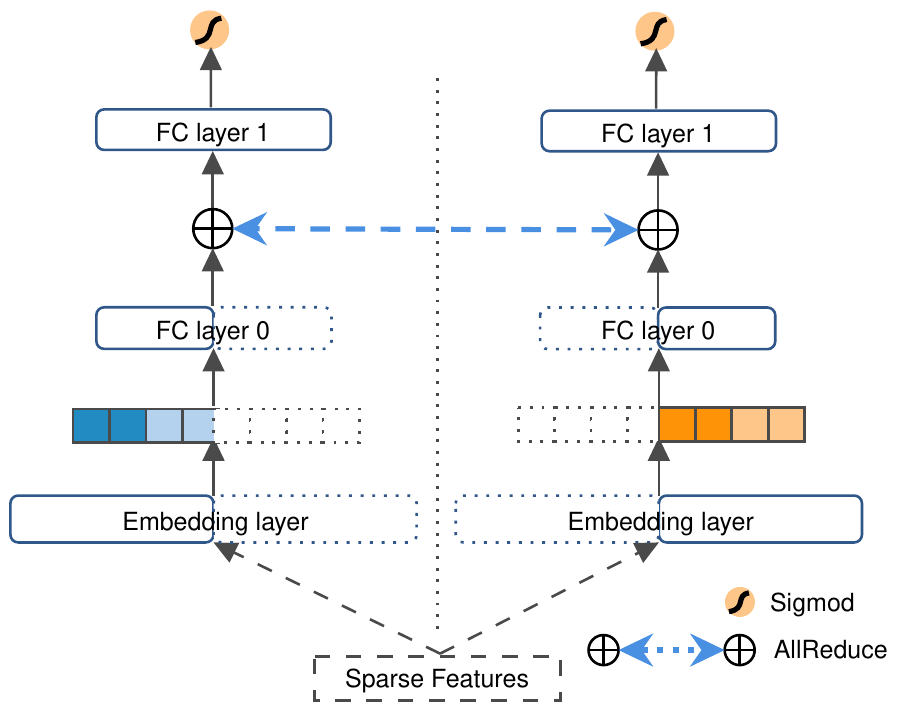}
	\caption{The architecture of DNN with 2 FC layers of PS-based strategy(left) and DES strategy(right)}
	\label{fig:dnn_ori}
\end{figure}

Recommender systems use DNN to learn high-order feature interactions. The features are usually categorical and grouped in fields. A DNN starts from an embedding layer which compresses the latent vectors into dense embedding vectors by fields, and is usually followed by multiple fully-connected layers as shown in Figure~\ref{fig:dnn_ori}.

Like FM, in DNNs, the majority of weights are from the embedding layer and the first FC layer:
\begin{equation}
\begin{aligned}
    F_{dnn}=\boldsymbol{V^T} \boldsymbol{W}
    \label{eq:dnn_ori}
\end{aligned}
\end{equation}
$\boldsymbol{V}$ denotes the concated output of the embedding layer and $\boldsymbol{W}$ denotes the weights of the first FC layer.

\begin{figure}[htbp]
	\centering
	\includegraphics[width=0.5\textwidth]{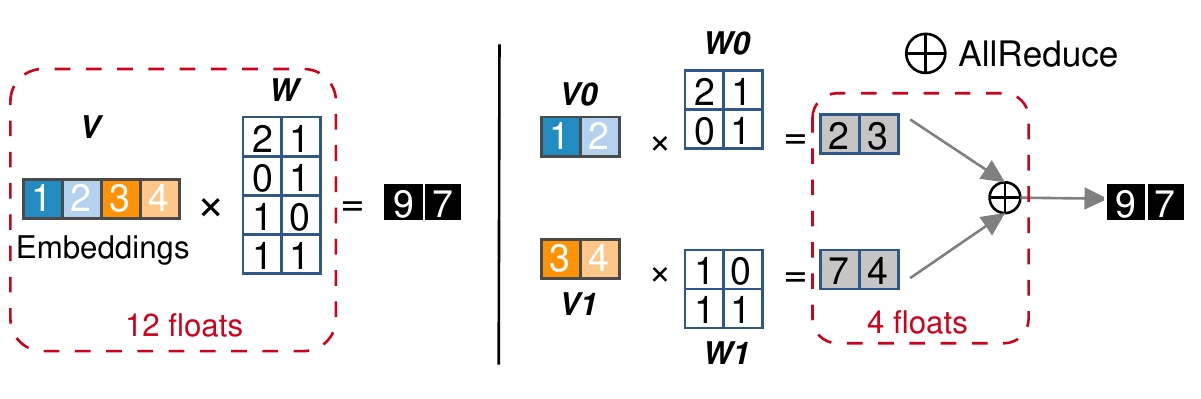}
	\caption{The blocked matrix multiplication in DNN using DES strategy(right).}
	\label{fig:dnn_impl2}
\end{figure}

Using DES, we split $\boldsymbol{V}$ and $\boldsymbol{W}$ into $N$ partitions over the fields dimension, and use blocked matrix multiplication (Figure~\ref{fig:dnn_impl2}), which is similar to the method proposed by Gholami  et al.~\cite{Gholami:2017:IMB}. Our strategy differs in splitting: we divide $\boldsymbol{V}$ and $\boldsymbol{W}$ in the same dimension to ensure that the computation and weights do not overlap in different parts:
\begin{equation}
\begin{aligned}
    \boldsymbol{V^T} \boldsymbol{W}&=\left[\begin{array}
    {c|c|c}
    {V_1^T} & {\hdots} & {V_N^T}\end{array}\right] \times \left[\begin{array}{c}{W_1} \\ \hline {\vdots} \\ \hline W_N \end{array}\right] \\
    \\
    &=\left[V_{1}^T W_{1}+{\hdots}+V_{N}^T W_{N}\right]
    \label{eq:dnn_bmm}
\end{aligned}
\end{equation}
Hence we get the $N\text{-partitions}$ of $\mathbf{V}$ and $\mathbf{W}$: $\mathbf{V}=\bigcup_{i=1}^{N} V_{i}$, $\mathbf{W}=\bigcup_{i=1}^{N} W_{i}$, where $W_{i}$ and $V_{i}$ denote the subset of $\mathbf{V}$ and $\mathbf{W}$  co-located with the $i$-th process respectively.

Considering that the embedding layer will aggregate the latent vectors by fields before concatenating them, we store the latent vectors of the same field on the same process to avoid unnecessary weights exchange. In this way, we also avoid communication during the back-propagation phase.

Using this $N$-partition we can define the local operator as follows:
$$\mathcal{M}_1(V_i,W_i)=V_i^T  W_i$$
The distributed equivalent substitution $f_{i}^{dnn}$ of $F_{dnn}$ is hence defined as:
\begin{equation}
\begin{aligned}
    f_{i}^{dnn} = \mathcal{F}_{dnn}(\mathcal{M}_i) =\oplus \mathcal{M}_1\left(V_{i}, W_{i}\right)
    \label{eq:dnn_des}
\end{aligned}
\end{equation}

In mesh-based strategy, each worker needs to lookup $\frac{N-1}{N}$ of $\boldsymbol{V}$ and $\boldsymbol{W}$ by keys(unsigned int64) from the hash tables co-located with other workers. The total data size to transfer for each worker is:
$$Q_{dnn}^{Mesh}=\frac{(N-1)}{N} \left(S_{f}+S_{V}+S_{W}\right)$$
$S_f$, $S_V$ and $S_W$ denote the size of feature keys, $\boldsymbol{V}$ and $\boldsymbol{W}$ per batch respectively. Compared to mesh-based strategy, DNN using DES only requires all workers to exchange $\mathcal{M}_1$ among each other (Figure~\ref{fig:dnn_ori}):
$$Q_{dnn}^{DES}=2\frac{(N-1)}{N} S_{\mathcal{M}_1}$$
 

The communication-saving ratio for DNN is:
$$\mathcal{R}_{dnn}=1-\frac{Q_{dnn}^{DES}}{Q_{dnn}^{Mesh}}=1- \frac{2S_{\mathcal{M}_1}}{S_k+S_V+S_W}$$

\begin{table}[h]
\caption{The number of unique features and communication-saving ratio of different models using a 4-node cluster.}
\label{table:r_x}
\centering
\begin{tabular}{llccr}
\toprule
batch & uniq\_feats & $\mathcal{R}_{lr}(\%)$ & $\mathcal{R}_{fm(2)}(\%)$ & $\mathcal{R}_{dnn}(\%)$\\
\midrule
512  & 	147,664 & 99.769	\% & 	99.376	\% & 	90.310	\% \\
1024 & 	257,757 & 99.735	\% & 	99.285	\% & 	86.226	\% \\
2048 & 	448,814 & 99.696	\% & 	99.179	\% & 	81.658	\% \\
4096 & 	789,511 & 99.654	\% & 	99.066	\% & 	77.015	\% \\
8192 & 	1,389,353 & 99.607	\% & 	98.939	\% & 	72.264	\% \\
\bottomrule
\end{tabular}
\end{table}

Using DES does not increase the computation compared to PS/mesh-based strategy, and often leads to smaller computation load. Table~\ref{table:r_x} shows the number of unique features per batch as well as the communication-saving ratio for three models with different batch sizes on a real-world recommender systems. The communication costs when using DES are reduced from 72.26\% (with a batch size of 8192) to 99.77\% (with a batch size of 512) compared to mesh-based strategy.

Our analysis here only include the communication cost for transferring the sparse weights. In fact, for most recommender systems, state-of-the-art stateful optimizer such as FTRL~\cite{McMahan:2013:FTRL}, AdaGrad~\cite{Duchi:2011:ADA} and Adam ~\cite{Diederik:2014:ADAM} require saving and transferring the corresponding state variables as well as the sparse weights. When using DES strategy, these variables are kept local, which will reduce even more communication cost.

\textbf{Extending to General Models:} Previous analysis shows that we can apply DES to several state-of-the-art models for recommender systems. We think this is not a coincidence. To generalize our observations for the above models, we claim that for any DLRM, as long as the computational equivalent substitution of the weights-rich layers do not surpass linear complexity, we can apply DES strategy. FM~\cite{Rendle:2010:FM} is the work that inspired us on finding linear substitution to operators. The linear complexity is $O(M)$ where $M$ is the size of the feature parameters. Since DES splits an $M$-dimension feature vector to $N$ part where $k*N=M$, $k$ is a constant, and $N$ is the number of DES worker processes. We use $O(M)$  to represent this. We have a simple rule to judge whether it has linear complexity or not: if the computation process of weights-rich layer satisfies \textbf{the Commutative Law and Associative Law}, we can apply DES strategy to help reduce the communication cost in forward phase and eliminate the gradient aggregation in backward phase. As a further proof, we confirm that DES can be applied to many mainstream DLRMs as shown in Table~\ref{table:general}.

\begin{figure}[htbp]
	\centering
	\includegraphics[width=0.40\textwidth]{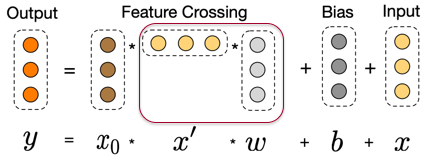}
\caption{Cross layer in DCN.}
\label{fig:dcn}
\end{figure}

\begin{table}[h]
\caption{Universal Generality on Mainstream DLRMs.}
\label{table:general}
\vskip 0.15in
\begin{center}
\begin{small}
\begin{sc}
\begin{tabular}{llr}
\toprule
Model & Weight-rich layer & DES Policy \\
\midrule
PNN~\cite{qu:2016:PNN}    & Product & same as FM \\
DCN~\cite{Ruoxi:2017:DCN} & Embedding, Cross  & shown in Fig.~\ref{fig:dcn} \\
AutoInt~\cite{Song:2019:AUTOINT}    & Product & same as DNN \\
xDeepFM~\cite{Lian:2018:XDEEPFM}    & FM, Embedding & same as DeepFM \\
DIEN~\cite{zhou:2018:DIEN}    & Embedding & same as DNN \\
FLEN~\cite{chen:2019:FLEN}    & FwBI, Embedding & same as DeepFM \\

\bottomrule
\end{tabular}
\end{sc}
\end{small}
\end{center}
\vskip -0.1in
\end{table}
\label{gr:dcn_explain}There are some differences between DCN and other models which are worth explaining separately. DCN uses the same DES policy on the embedding layer as DNN. The equation for using DES on the cross layer\footnote{For more details, Please refer to Section 2.2 and Figure 2 in the DCN paper~\cite{Ruoxi:2017:DCN}} is shown in Figure~\ref{fig:dcn}.

Where $w$ is the weights of cross-layer. On DES, we split the $w$ on the long dimension $d$($d$ has the same meaning as in the DCN paper) and saved on each workers seperatly, then the equation of DES for cross layer $y$ would be: 
$${y'}= x_0 * \oplus(x'_{local} * w_{local}) + b + x$$
Where we use the law of combination and compute firstly the 
$(x'_{local} * w_{local})$ on each workers which the result is only a scalar, so the following AllReduce on a scalar cross workers will require less run time than pulling the whole $w$ from remote PS. 

\section{System Implementation}
\label{sec:sys_impl}
We choose TensorFlow as the backend for our training framework due to its flexibility and natural distributed-friendliness. More specifically, we implement our system by enhancing TensorFlow in the following two aspects: large-scale sparse features and dynamic hash table.

\textbf{Large-scale Sparse Features:} As mentioned earlier, an industrial streaming recommender system may have hundreds of billions of dynamic features. Given the embedding size $d=8$ with $float32$, the feature weights require 3.2TB of memory at least. Table~\ref{table:r_x} shows that for a single iteration, weights update on unique features is sparse. To achieve constant cost data access/update and get over the memory constraint of a single node, we use distributed hash table. We use a simple method to distribute weights: In a cluster with $N$ nodes, the $i$-th node will hold all the weights that are corresponding with feature field IDs $f$ where $i=f\mod N$. There are other methods that could achieve better load balancing, but we found this simple method works fine in our case.

\begin{figure}[htbp]
	\centering
	\includegraphics[width=0.28\textwidth]{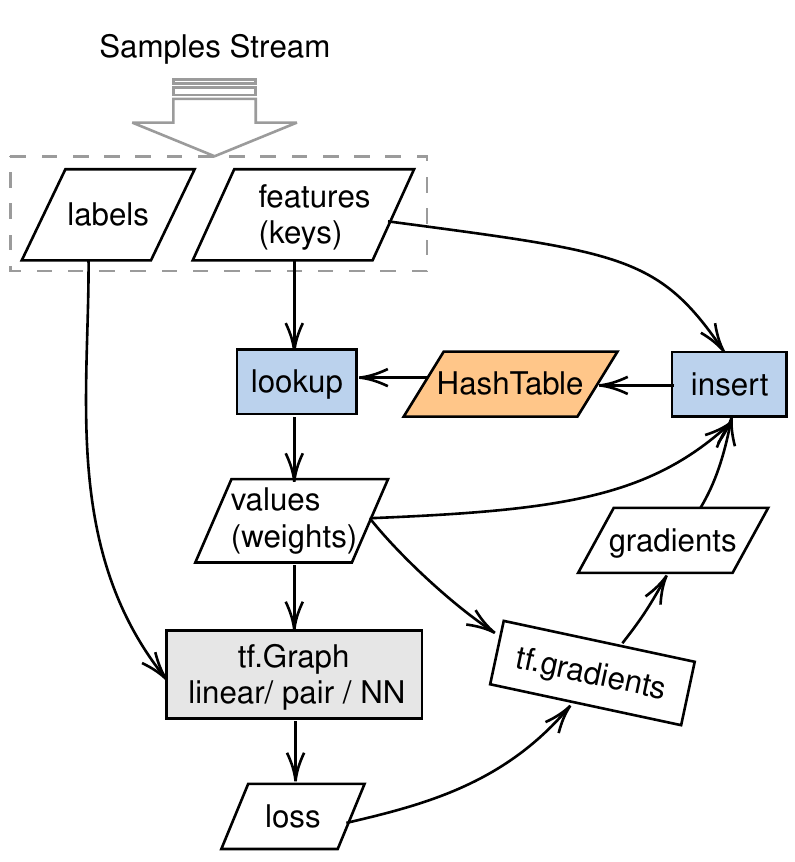}
	\caption{Data flow chart with our enhanced TensorFlow.(The two operators of \textit{lookup} and \textit{insert} isolate the sparse domain.)}
	\label{fig:flow}
\end{figure}

\textbf{Dynamic Hash Table:} In DES strategy, there are three places we operate on hash tables: given a feature ID in a batch of input samples, we \textit{lookup} the corresponding weight; when a new feature ID is given as the key, we \textit{insert} the initialized weight into the hash table; given the gradient of a weight, we apply it locally, and then \textit{update} the hash table with the new weight. To achieve this, we provide a modified dynamic hash table implementation in TensorFlow with key operations adapted to our needs (Figure~\ref{fig:flow}). Compared to alternative design choices, this implementation makes use of as many existing TensorFlow features as possible but only introduces hash table operations during batch building and optimizer phase. Because after the \textit{lookup}, the sparse weights are reformed into dense tensors and are fully compatible with the native training pipeline of TensorFlow.

\section{Experiments And Analysis}
\label{sec:exp}
\textbf{Hardware:} We ran all experiments in this paper on a testing cluster which has four LINUX servers with each consisting of 2 hyperthreaded 24-core Intel Xeon E5-2670v3(2.3GHz) CPUs, 128 GB of host memory, and one Intel Ethernet Controller 10-Gigabit X540-AT2 without RDMA support. 

\noindent\textbf{Software:} Our DES framework is based on an enhanced version of TensorFlow 1.13.1 and a standard OpenMPI with version 4.0.1. Considering that mesh-based frameworks is a special form of PS-based and usually has less communication cost than original PS-based frameworks, we use mesh-based strategy for comparison. The mesh-based strategy we compare with is implemented using a popular open-source framework: DiFacto~\cite{Li:2016:DDF}.


\noindent\textbf{Dataset:} In order to verify the performance of DES in real industrial context, we evaluate our framework on the following two datasets.

\textbf{1) Criteo Dataset:} Criteo dataset\footnote{\url{http://labs.criteo.com/downloads/2014-kaggle-displayadvertising-challenge-dataset/}} includes 45 million users' click records with 13 continuous features and 26 categorical features. We use 95\% for training and the rest 5\% for testing.

\textbf{2) Company* Dataset:} We extract a continuous segment of samples from a recommender system in use internally. On average, each sample contains 950 unique feature values. The total number of samples is 10,809,440. It is stored in a remote sample server.
\noindent\textbf{Parameter Settings:} We set DiFacto to run one worker process on each server, the batch size is 4,096, and the number of concurrency threads is 24. Correspondingly, the parameters of $intra\_op$ $\_parallelism\_threads$ and $inter\_op\_parallelism\_threads$ for DES on TensorFlow are both set to 24,  the batch size on DES is set to 4096 when testing AUC~\@. Since for DES, all workers train samples from the same batch synchronously in parallel, when testing communication ratio, we set the batch size to 16384 (for $N$=4) to guarantee a fair comparison. We train all models with the same optimizer setting: FTRL for order-1 components, AdgaGrad or Adam for both Embedding and DNN components.

\noindent\textbf{Evaluation Metrics:} We use two evaluation metrics in our experiments: \textbf{AUC} (Area Under ROC) and \textbf{Logloss} (cross entropy).

\noindent\textbf{Performance Summary} We compare our framework to mesh-based implementation on three different widely-adopted models in mainstream recommender systems: LR, W\&D, and DeepFM~\@. As DES uses synchronous training, it will not be affected by the stale gradients problem~\cite{dutta:2018:STALEGRAD} and can achieve better AUC in smaller number of iterations with an order of magnitude smaller communication cost.

\begin{figure}[h]
	\label{fig:auc}
	\centering
	\includegraphics[width=0.4\textwidth]{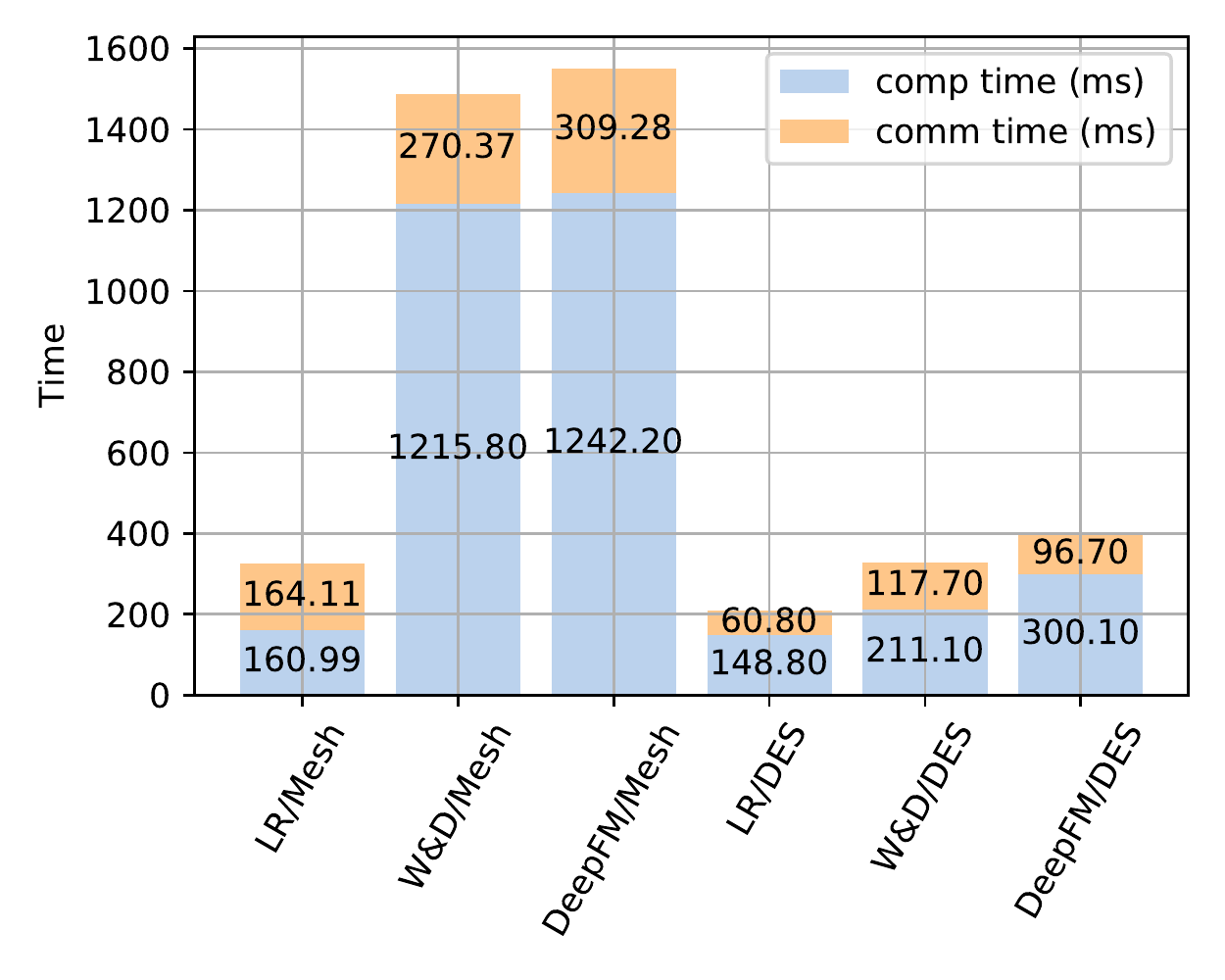}
	\caption{Per-iteration computation and communication time for three models.}
	\label{fig:comm}
\end{figure}

\textbf{Computation vs. Communication Time:} Figure~\ref{fig:comm} shows that in all experiments, DiFacto framework needs to spend more time on both computation and communication. The absolute total network communication time using DiFacto framework is 2.7x, 2.3x, and 3.2x larger for LR, W\&D, and DeepFM respectively, than using DES~\@. The saving on communication time comes from the smaller amount of intermediate results sent among workers during the forward phase and the elimination of gradient aggregation during the backward phase. The saving on computation time comes from the reduced time complexity of computational equivalent substitution as well as several optimizations we have put in our DES framework.

\textbf{Throughput:} Table~\ref{table:throughput} compares the throughput of DES and DiFacto. For deep models with high-order components (W\&D and DeepFM), DES has more advantages. It indicates larger benefits when applying DES to future DLRMs.

\begin{table}[h]
\caption{Throughput of DES and PS on three models.}
\label{table:throughput}
\vskip 0.15in
\begin{center}
\begin{small}
\begin{sc}
\begin{tabular}{lccc}
\toprule
model & \multicolumn{2}{c}{Throughput (samples/sec)} & improvement \\
\midrule
      & PS & DES & \\
LR    & 50396.8 & 78205.3 & 1.55x \\
W\&D  & 11023.9 & 49837.3 & 4.52x \\
DeepFM & 10560.1 & 41295.5 & 3.91x \\
\bottomrule
\end{tabular}
\end{sc}
\end{small}
\end{center}
\vskip -0.1in
\end{table}

\begin{table}[h]
\caption{Average AUC for three models after a 7-day training session on Company* Dataset, DNN 3-layers.}
\label{table:auc}
\vskip 0.15in
\begin{center}
\begin{small}
\begin{sc}
\begin{tabular}{lccr}
\toprule
policy & min & max & avg \\
\midrule
PS & 0.7909 & 0.8315 & 0.8134 \\
DES & 0.8038 & 0.8407 & 0.8244 \\
\bottomrule
\end{tabular}
\end{sc}
\end{small}
\end{center}
\vskip -0.1in
\end{table}

Table~\ref{table:auc} shows that during long-term online training, when consuming the same amount of samples with similar distribution, DES shows better average AUC for all three models. One possible explanation for this is that with DES, the training is in synchronous mode, which usually leads to better and faster convergence compared to asynchronous mode~\cite{dutta:2018:STALEGRAD}. The reason we care about small amount AUC increase is that in several real-world applications we run internally, even $0.1\%$ increase in AUC will have a 5x amplification ($0.5\%$ increase) when transferred to final CTR~\@.

\begin{table}[h]
\caption{Average AUC and log loss for three models using PS (async training) and DES (sync training) with TensorFlow after a one epoch training session on Criteo Dataset.}
\label{table:epoch}
\vskip 0.15in
\begin{center}
\begin{small}
\begin{sc}
\begin{tabular}{lcccr}
\toprule
model & PS &  & DES  &  \\
\midrule
      & AUC & LogLoss & AUC & LogLoss \\ 
W\&D  & 0.7819 & 0.4765 & 0.7978 & 0.4528 \\
DeepFM & 0.7923 & 0.4674 & 0.8005 & 0.4505 \\
FM     & 0.7922 & 0.4666 & 0.8007 & 0.4506 \\
\bottomrule
\end{tabular}
\end{sc}
\end{small}
\end{center}
\vskip -0.1in
\end{table}

Table~\ref{table:epoch} shows the AUC and log loss for three models using PS-mode asynchronous training and DES-mode fully-synchronous training on TensorFlow respectively\footnote{We use FTRL optimizer for LR model (Wide component), and Adam optimizer for the other two models.}. The batch size is set to 2,048. As the convergence curve does not change much later, we only show the results after the first epoch. For PS-mode, we use 15 parameter servers (with 10GB memory) and 20 workers (with 5GB memory); for DES-mode, we use 15 workers (with 10GB memory). The one-epoch results show that DES have reached higher AUC on all three models (boosts are from 0.84\% to 1.6\%) even at very early stage during the training.

\section{Conclusions and Future Works}
\label{sec:conclusion}
We propose a novel framework for models with large-scale sparse dynamic features in streaming recommender systems. Our framework achieves efficient synchronous distributed training due to its core component: Distributed Equivalent Substitution (DES) algorithm. We take advantage of the observation that for all models in recommender systems, the first one or few weights-rich layers only participate in straightforward computation, and can be replaced by a group of distributed operators that form a computationally equivalent substitution. Using DES, the intermediate information needed to transfer between workers during the forward phase has been reduced, the AllReduce on gradients between workers during the backward phase has been eliminated. The application of DES on popular DLRMs such as FM, DNN, Wide\&Deep, and DeepFM shows the universal generality of our algorithm. Experiments on a public dataset and an internal dataset that compare our implementation with a popular PS-based implementation show that our framework achieves up to 68.7\% communication savings and higher AUC~\@.

\textbf{Future Works:} We have shown in section~\ref{sec:exp} that our current implementation of DES is bounded by computation. So the natural next step is to transfer the computation of current bottleneck operators such as hash table to GPU and to improve the existing kernel implementations. We have also started the initial work to apply DES to more models commonly used in industry such as DCN~\cite{Ruoxi:2017:DCN} and DIN~\cite{Zhou:2018:DIN}.

\section{Acknowledgement}
We appreciate the technical assistance, advice and machine access from colleagues at Tencent: Chaonan Guo and Fei Sun. We also thank the anonymous reviewers for their insightful comments and suggestions.
 
\bibliographystyle{ACM-Reference-Format}
\bibliography{des}

\end{document}